# What Is the Point of Equality in Machine Learning Fairness? Beyond Equality of Opportunity

*This article has been accepted for publication in the ACM Journal on Responsible Computing and will appear in a forthcoming issue. Please cite the final published version.

Youjin Kong

Department of Philosophy, Artificial Intelligence Institute, Institute for Gender Research, and American Studies Institute, Seoul National University, Seoul, South Korea, youjinkong@snu.ac.kr

**Abstract**: Fairness in machine learning (ML) has become a rapidly growing area of research. But why, in the first place, is unfairness in ML wrong? And why should we care about improving fairness? Most fair-ML research implicitly appeals to distributive equality: the idea that desirable benefits and goods, such as opportunities (e.g., Barocas et al., 2023), should be *equally distributed* across society. Unfair ML models, then, are seen as wrong because they unequally distribute such benefits. This paper argues that this exclusive focus on distributive equality offers an incomplete and potentially misleading ethical foundation, especially in the context of text- and image-generation models. Grounding ML fairness in egalitarianism—the view that equality is a fundamental moral and social ideal—requires challenging structural inequality: systematic, institutional, and durable arrangements that privilege some groups while disadvantaging others. Structural inequality manifests through ML systems in two primary forms: allocative harms (e.g., economic loss) and representational harms (e.g., stereotypes, erasure). While distributive equality helps address allocative harms, it fails to explain why representational harms are wrong—that is, why it is wrong for ML systems to reinforce social hierarchies that stratify people into superior and inferior groups—and why ML systems should aim to foster a society where people *relate as equals* (i.e., relational equality). To address these limitations, the paper proposes a novel *multifaceted egalitarian framework* for ML fairness that integrates both distributive and relational notions of equality. Drawing on critical social and political philosophy, including the work of Anderson, Young, and Fraser, this framework offers a more comprehensive ethical foundation for tackling the full spectrum of harms perpetuated by ML systems. The paper also outlines practical pathways for implementing the framework across the entire ML pipeline.

**CCS CONCEPTS** • Social and professional topics → User characteristics; Socio-technical systems; • Computing methodologies → Machine learning • Applied computing → Arts and humanities

**Additional Keywords and Phrases:** Machine Learning Fairness, Normative Foundation, Egalitarianism, Distributive Equality (DE), Equality of Opportunity (EOP), Relational Equality (RE), Allocative Harm, Representational Harm

## 1 Introduction: The Normative Turn in ML Fairness

The field of fairness in machine learning (ML) has traditionally focused on developing *statistical* metrics for measuring the (un)fairness of ML models. With over 20 definitions proposed [72], these metrics often fall into categories like individual fairness versus group fairness [13, 34], and encompasses criteria like demographic parity, error rate parity, and calibration [8, ch.3]. A significant portion of the fair-ML literature debates which definition best captures fairness, essentially arguing which metrics most effectively measure an ML model's fairness. These metrics address the "what" and "how" of ML fairness: identifying what bias exists within models and devising technical methods for debiasing them.

Recently, however, there has been a growing emphasis on establishing *normative* grounds for ML fairness, which involves providing ethical justifications. This shift seeks to address the "why" question: why bias in ML is wrong and should be opposed (negative justification), and why we should strive for fairness in ML as a normative ideal—something we believe is right and should be pursued (positive justification).[1]

Why is it crucial to establish normative grounds for ML fairness? Jacobs and Wallach [50] offer key insights. First, fairness is not something we can directly observe and measure. It can only be inferred through observable properties, such as statistical metrics mentioned earlier, that we assume to be linked to fairness (these metrics function as "operationalizations" of fairness). Second, fairness is a contested concept that depends on theoretical foundations of justice, which can sometimes contradict each other. These two factors, Jacobs and Wallach argue, lie at the root of debates surrounding ML fairness definitions. While these debates often focus on mathematical aspects (finding better statistical metrics), the core issue is actually the different normative understandings of fairness.

[1] In this paper, I use the term "normative" to encompass both moral-ethical and social-political values, since key notions in fair ML, such as fairness, equality, and justice, are relevant in both spheres.

For example, consider the debate surrounding the COMPAS recidivism prediction algorithm. ProPublica [5] revealed that this algorithm tended to mark Black defendants as higher risk, falsely labeling them as future criminals twice more often than their white counterparts. When they called the algorithm biased and unfair, they used "error rate parity" (equal false positive rates across races) as the fairness metric. Northpointe, the developer company, countered that the COMPAS algorithm was fair because it exhibited "predictive parity" (also known as "calibration"): "the probability of recidivism among Black defendants scored as high risk was the same as the probability of recidivism among white defendants as high risk" [50, p.383].

On the surface, this appears as a disagreement about how we measure fairness of a model: which observable characteristics, such as error rates and predictive values, are better metrics of fairness. However, it reflects deeper normative clashes about equality and justice. ProPublica highlights the unequal burden of misclassification, arguing that a fair system should not disproportionately label Black defendants as high risk, leading to harsher sentences. This perspective is grounded in the fundamental normative assumption that burden must be distributed equally across members of society. It is morally unjustifiable that one group bears more burden than another due to their racial identity, which is an arbitrary and irrelevant factor when it comes to justifying the inequality in burden distribution. Northpointe takes a different stance. For them, fairness lies in treating individuals with similar scores similarly. If two individuals are scored by the model as having the same risk but one is significantly more likely to recidivate than the other, that would be problematic. However, as long as the risk score reflects the actual chance of recidivism, they see no discrimination. In other words, the only type of parity that matters in characterizing an ML model as "fair" is that of predictive values. Other kinds of parity and disparity, such as those of error rates, do not affect the normative state of the model.

This debate exemplifies how fairness metrics are actually grounded in distinct normative theories. It is not only about choosing a metric, but more about what sorts of inequality are morally unjustifiable, and what forms of equality ML systems should strive to achieve. Different answers to these questions reflect different ethical and social-political understandings, namely, normative frameworks for fairness. By unpacking these normative grounds, we move beyond debates regarding how to measure and calculate fairness, and delve into the ethical core of ML fairness: what kind of equality constitutes ML fairness.

# 2 Distributive Equality and Equality of Opportunity: Prevailing Normative Frameworks for Fair ML

Given this emphasis on equality, egalitarianism—a theory that views equality as a moral and social ideal—emerges as the leading approach for justifying fairness in ML. Anchoring ML fairness in egalitarianism indicates that unfair ML models are seen as wrong because they create or amplify *inequality*. As a moral and social wrong, inequality in the fair-ML context typically refers to *discrimination*, that is, unequal treatment of individuals or groups based on morally arbitrary factors such as race, gender, and class. Conversely, pursuing fair ML becomes a moral and social imperative, as it corresponds to the normative goal of promoting *equality*.

Within egalitarianism, two main conceptions of equality exist. Distributive egalitarianism, or the **distributive conception of equality (DE)**, focuses on the equal distribution of goods and benefits. Different DE theories propose different things to distribute equally ("equality of what?"), such as equal opportunities, resources, welfare, capabilities, and fortune/luck. Relational egalitarianism, or the **relational conception of equality (RE)**, which I will discuss later, focuses on fostering equal social relations. Reuben Binns (2018), one of the first works that applied egalitarianism to fair ML, embraces both DE and RE: "Broadly speaking, egalitarianism is the idea that people should be *treated equally*, and (sometimes) that certain valuable things should be *equally distributed*" [13, p.5; emphases added].

However, most fair-ML literature seeking normative grounds draws on DE [48, 64-66, 89], particularly the concept of **equality of opportunity (EOP)** [21, 47, 49, 56]. EOP argues that everyone should have an equal chance to compete for desired outcomes (e.g., loans, jobs, and college admissions). As long as opportunities are equally distributed, the unequal distribution of desired resources or positions is considered acceptable and also justified [38, sec.1-2].[2] With a level playing field established, EOP allows ML models to justifiably assist decision-making processes that lead to inequality in outcomes [47, pp.1-2].

Based on their comprehensive analysis of fair-ML research, Solon Barocas, Moritz Hardt, and Arvind Narayanan (2023) in their book *Fairness and Machine Learning: Limitations and Opportunities* [8; hereafter referred to as BHN] also argue for DE, particularly EOP, as the normative foundation for ML fairness. I take their argument to hinge on both negative and positive justifications.

At the core of BHN's negative justification for EOP lies the concern that unfair ML models contribute to ongoing *discrimination*. As they note, ML models "don't operate in a vacuum; they are adopted in societies that already have many types of discrimination intertwined with systems of oppression such as racism" [BHN p.221]. BHN's account of discrimination closely aligns with Iris Marion Young's (1990) foundational theory of structural injustice, although they do not explicitly reference her work. What does it mean for injustice to be "structural"? Young notes that the widespread and enduring injustices experienced by some social groups are not the result of a few bad actors or isolated policies, but arise from the routine operation of everyday social processes [90, ch.2, 92, ch.2]. This account mirrors BHN's own usage, and they later explicitly analyze the concept of structural discrimination [BHN ch.8]. Accordingly, this paper uses the terms "structural injustice," "structural discrimination," and "structural inequality" interchangeably, as they refer to the same phenomenon: the systematic, institutional, and durable processes through which social hierarchies are maintained and certain groups are persistently disadvantaged.

[2] Among different notions of distributive equality, this paper focuses on equal opportunity and the combined notion of equal resources/outcomes. While "equal resources" and "equal outcomes" are not exactly the same concepts, the paper uses them interchangeably when discussing the distribution of *resources* resulting as the *outcome* of ML-based decision-making.

Young's theory highlights these three interrelated characteristics of structural discrimination. First, structural discrimination is *systematic*. Echoing Young, political philosopher Elizabeth Anderson argues that isolated instances of unfairness, such as a single discriminatory act against an individual for having green eyes, do not constitute structural inequality. Instead, stable and routine systems in society create and sustain hierarchies, rendering some social groups superior to others based on group identities such as ethnicity, language, citizenship status, and sexuality [3, p.43]. BHN similarly note that structural inequality is not merely "different treatment" but treatment that "systematically imposes a disadvantage on one social group relative to others … according to characteristics like race, gender, or disability" [BHN pp.83-85]. Second, it is *institutional*. "Institutions" here include not only formal structures like legal and governmental systems but also informal and "institutionalized" ones, such as market mechanisms, media stereotypes, and social norms that shape behavior and interactions. Third, it is *durable*. Anderson points out structural inequalities are "reproduced over time by the social arrangements that embody them," including laws, norms, and habits [3, p.43].

The systematic, institutional, and durable nature of structural discrimination explains why dismantling it is so challenging. This discrimination is perpetuated through normal(ized) societal process, including ML systems. As a key institution constituting contemporary society, ML systems do not merely reflect existing societal inequalities but actively sustain and amplify them. For this reason, ML systems are not merely "technical" systems but are better understood as "sociotechnical" systems. Drawing on Young's concept of structural injustice, Kasirzadeh [52, sec.3] highlights how these sociotechnical systems make marginalized social groups vulnerable to further injustice, even in the absence of a clear individual (e.g., tech giants, their employees, or crowdworkers) to blame and hold liable [92, ch.2].

Understanding ML unfairness through the lens of structural discrimination also suggests that it cannot be diagnosed or addressed solely within the ML model itself. For example, BHN discuss a case study of the gender earnings gap on Uber [26]. The study attributes the gap to experience, speed, and location and concludes that it is therefore not due to algorithmic and customer discrimination. However, BHN argue that this conclusion is based on a narrow understanding of discrimination. For instance, the study does not consider critical factors like women drivers leaving the platform due to customer harassment (experience), facing harsher social penalties for driving fast ("aggressive women") compared to their male counterparts (speed), and safety concerns leading to less lucrative routes (location). These factors are "example[s] of the greater burden that society places on women. In other words, Uber operates in a society in which women face discrimination and have unequal access to opportunities, and the platform perpetuates those differences in the form of a pay gap" [BHN pp.223-24].

This case illustrates how ML systems can reinforce structural discrimination. Stereotypes about how women should behave, women's vulnerability to physical and sexual violence influencing their decisions, and the resulting patterns embedded in Uber's operations collectively lead to women earning less than men. That is, ML unfairness is less about unequal treatment per se but more about how ML systems, as a social institution itself, systematically sustain and durably reproduce structural inequalities—here, manifesting as a gender pay gap in ridesharing platforms.

Building on this negative justification, BHN propose EOP as the positive normative ideal for fair ML. They present EOP as a widely recognized goal of limiting discrimination [BHN p.88] and a common conceptualization of fairness [p.110]. According to their argument, an ML model is fair and morally right when it fosters "progress toward enabling equality of opportunity" [p.229], positioning EOP as the ethical and social aim of ML fairness. BHN further distinguish three views of EOP, each tied to statistical fairness metrics in ML [ch.4], which I will analyze alongside related concepts from the philosophy literature (see Table 1).

**Table 1: Three views of equality of opportunity discussed by Barocas, Hardt, and Narayanan (2023) and their philosophical connections**

| EOP View | Statistical Criteria | Philosophical Connection | Structural Inequality | Scale of Intervention |
|---|---|---|---|---|
| **Narrow** | Calibration | Meritocratic EOP | Not Considered | ML Model Design |
| **Middle** | Error Rate Parity | Equal Distribution of Burden | Acknowledged | ML Model Design |
| **Broad** | Demographic Parity | Rawls's Fair EOP | Addressed through Reform | Societal Institutions |

According to the narrow view, if a model ensures that “people who are similarly qualified for an opportunity have similar chances of obtaining it,” it achieves EOP [BHN p.94]. This view aligns with the statistical criterion of calibration: those with similar scores on task-relevant features have similar rates of positive outcomes. In the philosophical literature on EOP, it connects to a thinner conception of EOP such as meritocracy, where selection is based on qualifications judged relevant to the task at hand [38, sec.4]. A problem of this narrow, meritocratic EOP is that it does not dictate how we should define “similarly qualified” or “similar scores” [BHN p.90]. Intervening only at the decision-making moment, the narrow view does not consider how structural inequality, such as societal biases, can shape people’s qualifications in the first place.

To tackle this issue, the broad view of EOP directs attention to the overarching organization of society, aiming to ensure that “people of equal ability and ambition are able to realize their potential equally well” [BHN p.94]. This perspective reflects John Rawls’s philosophical concept of “fair equality of opportunity,” which emphasizes two key points. First, positions and offices should be open to all, adhering to the “careers open to talents” principle. Second, everyone should have a fair chance to attain those positions: the prospect of success for individuals with the same level of talents, skills, and willingness to use them should not be influenced by their socioeconomic background [79, sec.11-12]. While BHN associate the broad view of EOP with demographic parity (i.e., a statistical criterion requiring equal acceptance rates across demographic groups), they note that the broad view primarily concerns the design of society’s basic institutions, such as education, rather than the specifics of ML model design [BHN p.100]. This suggests that the broad view of EOP might not be the most suitable normative ground for ML fairness.

The middle view, which BHN endorse, bridges the gap between the narrow and broad views by discounting difference in qualification due to structural discrimination. Unlike the narrow view, it considers historical injustices that contribute to current qualification gaps. However, unlike the broad view, it addresses these issues at the level of ML model design and decision-making processes, making it more relevant to achieving fairness in ML applications [BHN pp.91-92]. The middle view aligns with the statistical standard of error rate parity. Disproportionately high error rates, such as those observed in the COMPAS recidivism prediction algorithm, serve as a quantitative proxy for the social burden experienced by marginalized groups. The middle view argues that equalizing error rates across groups promotes fairer opportunities by reducing this burden [pp.105-06].

To sum up, BHN’s argument can be restructured in two steps. (1) **Fairness = Nondiscrimination**; or conversely, Unfairness = Discrimination. Unfair ML systems are morally unjustifiable because they perpetuate structural discrimination and inequality. We should work to make ML fairer because it helps us make progress towards a nondiscriminatory society. (2) **Fairness as Nondiscrimination = Equality of Opportunity**. Here I would like to point out that all three views of EOP discussed by BHN aim to *distribute* benefits and burdens equally among members of society: whether “equal opportunity” is interpreted as equal chance of obtaining benefits such as a job based on the score/qualification (narrow view), as equal prospect of long-term success through, for example, equal access to education (broad view), or as discounting disadvantage due to structural inequality (middle view), they all share the assumption that ML systems should aim for equal distribution of such benefits or burdens. This focus on distribution underlies BHN’s conception of EOP as the normative ground for ML fairness.

I agree with the first part of the argument, but I disagree with the second part: I do not think that EOP is what makes fair ML morally right or is the normative ideal that fair-ML efforts should aim to achieve. While EOP can contribute to a nondiscriminatory society, EOP alone is insufficient. I argue that equal distribution of opportunity, or the interpretation of equality in terms of distribution in general, does not capture the full spectrum of harms caused by structural discrimination embedded in ML systems.

Before delving into this argument, it would be helpful to address the first part of BHN’s argument: the understanding of ML fairness and its normative goal as nondiscrimination. Some researchers may dissent from this notion, contending that not all aspects of fair ML involve structural discrimination. For example, they point to cases of “statistical bias” [29], such as a car insurance model falsely predicting that drivers of red cars are more likely to be involved in costly accidents [BHN p.38]. In such instances, achieving “statistical fairness” by removing bias is not related to structural discrimination in society.

To respond to this counterpoint, I make two clarifications. First, it is important to distinguish between debiasing and fairness. While much of the fair-ML literature has focused on debiasing, which addresses unfairness to some extent (thereby promoting “weak” fairness), a more robust and “strong” sense of fairness requires actively challenging structural discrimination and fostering a fairer society [62]. Second, even if some researchers do not embrace this “strong” conception of fairness, it is essential to consider the normative justification for fair ML—why bias in ML is morally wrong and why improving ML fairness is a moral imperative (see Section 1). Cases of statistical bias unrelated to structural inequality, such as the insurance example, do not present core normative issues. The primary problem is economic: the model fails to incorporate relevant information (e.g., driving behavior instead of car color), leading to flawed inductive reasoning and unnecessary costs for both red car owners and the company. The justification for fixing this model is to bring greater financial benefits for stakeholders by improving predictions, rather than because the model ethically fails the drivers or because the company has ethical obligations to redress such failure. Therefore, while addressing statistical biases could be practically important, it does not significantly contribute to the normative justification of fair ML. This paper proceeds on the premise that ML fairness, when unrelated to structural discrimination, offers only limited insights into the ethical foundations of fair ML.

# 3 Allocative and Representational Harms: Strengths and Limits of Distributive Equality

Kate Crawford's (2017) influential keynote [28], along with the subsequent co-authored paper [7], delineated two primary forms of harm resulting from biased systems. *Allocative harm* arises when ML systems withhold opportunities and resources from certain groups. *Representational harm*, on the other hand, occurs when ML systems reinforce the subordination of some groups based on their identities. I argue that while DE offers a compelling framework for addressing allocative harms, it falls short in capturing representational harms.

To begin with **allocative harms**, Shelby and colleagues [82] delineate two main avenues through which these harms occur. *Opportunity loss* arises when ML systems hinder certain groups' access to information or opportunities to pursue desired goods and positions. For instance, ad targeting systems may exclude racial minorities from viewing housing ads [4]. Even in the case where advertisers do not explicitly target their ads based on race or ethnicity (which is often illegal), biases might infiltrate datasets and model behavior, leading to the exclusion of minority groups from the targeted audience [BHN pp.208-10]. Another example is job search algorithms that favor men over women by making them more visible to potential employers [83]. Such biased algorithms can prevent qualified individuals from even being considered for opportunities.

*Economic loss*, another facet of allocative harm, refers to the negative financial consequences of unfair algorithms. While often intertwined with opportunity loss, economic loss directly relates to the financial harm caused by ML systems [82, p.730]. For example, a group of LGBTQ content creators sued YouTube, alleging that the platform's algorithms falsely flagged safe content containing keywords like "transgender," "lesbian," or "gay" as adult content, preventing them from earning ad revenue [10, 57]. Another study found that wealthier neighborhoods benefit more from the economy-sharing platforms like TaskRabbit and Uber [85]. This disparity often translates into racial inequalities, given the relation between poverty and race [BHN p.212].

In essence, allocative harms show how opportunities and resources are distributed through ML systems in a way that disadvantages certain groups based on morally irrelevant factors, such as race, gender, and sexuality. It is one of the avenues in which structural discrimination manifests through ML, namely, *unequal distribution*. DE, with its emphasis on *equal distribution* of opportunities and other resources, provides an ethical framework for challenging these harms. Notably, EOP directly addresses opportunity loss by ensuring everyone has a fair shot at achieving their goals. In the context of ML, this translates to designing models that avoid perpetuating biases that disadvantage certain groups in areas like access to housing ads or job opportunities. Economic loss can be tackled through other branches of DE, such as equality of resources and outcomes. The normative principle demanding equal distribution of resources can inform the development of models that consider the potential economic impact on different groups. Distributive egalitarianism thus provides a clear justification for why these harms are morally wrong and should be tackled. A normative foundation for fairness in ML needs to have DE as part of it, as I will further argue in Section 5.

However, focusing solely on DE as such a normative foundation fails to adequately capture the moral wrongness of **representational harms** and the normative ideal that we strive for by addressing them. To delve into this argument, let us explore a more comprehensive understanding of representational harm. Since Crawford and colleagues first called attention to representational harm in 2017 [7, 28], significant progress has made in categorizing its various forms [53, 54, 82, 88]. I will focus on the four categories of representational harm most frequently identified in these works: stereotyping, demeaning, erasing, and reifying. I will present their definitions and examples from the fair-ML literature while also reframing them through insights from philosophical literature.

Before discussing each category, it is important to clarify their ontological character (what these categories are) and their methodological role (how they function in analysis): these four categories serve as *criteria* for identifying whether ML models produce representational harms. While there might be no strict quantifiable metrics for assessing representational harms, these categories can provide a useful framework for guiding such evaluations. Drawing on Young's discussion of the "five faces of oppression"—where the presence of any of the five faces, such as exploitation and powerlessness, is sufficient to identify a social group as oppressed [90, pp.63-65]—I argue that the occurrence of stereotyping, demeaning, erasing, and/or reifying of social groups or group identities by an ML model is sufficient to determine that representational harm exists. A single instance of representational harm may fall into one or more of these categories, as Crawford initially suggested. Specific examples discussed below will illustrate how these categories overlap and interact in practice.

1) ML systems can perpetuate *stereotypes* about social groups. Consider translating the English sentences "She is a doctor" and "He is a nurse" into Turkish, a language with gender-neutral pronouns. The correct translations would be "O bir doktor" ("He/She/They is a doctor") and "O bir hemşire" ("He/She/They is a nurse"). However, translating these sentences back into English using translation engines often produces "He is a doctor" and "She is a nurse" [BHN p.11]. Similarly, text-to-image models might consistently depict "CEO" as male or generate sexualized images when prompted with terms like "lesbian" [46]. These outputs reinforce stereotypes and can have downstream impacts. For example, stereotypical images of CEOs as male might discourage young girls from pursuing such positions, while sexualized portrayals of lesbians perpetuate homophobic views.

One might argue that these models simply reflect existing societal realities—after all, doctors and CEOs are predominantly male, while nurses are predominantly female. Isn't the model merely mirroring the world as it is rather than stereotyping? This view, I argue, is rooted in what Beeghly [9] calls the "falsity hypothesis"—that stereotypes indicate false or inaccurate views of groups. Indeed, the fair-ML literature often defines stereotypes as "overgeneralized" or "oversimplified" beliefs [54, 88], implying that they are problematic primarily because they falsely generalize.

However, the harm of stereotypes extends beyond their truth or falsity. As Beeghly argues, "stereotypes are about power, and they exert power" [9, p.53]. Drawing on Patricia Hill Collins's concept of "controlling images" [25], Beeghly notes that stereotypes maintain social hierarchies and often promote domination [9, p.52]. When ML systems portray doctors and CEOs as male, they do more than reflect societal patterns; they actively reinforce gendered social hierarchies, propagating the expectation that positions of power and prestige are male-dominated.

2) *Demeaning* is similar to stereotyping but focuses more on how social groups are portrayed as lower in status and less deserving of respect [88, p.327]. For example, our research team found during exploratory research that large language models developed by major South Korean tech companies produced outputs demeaning South Asians. When prompted to generate a story involving terms like "unsolved murder case," "culprit," "a Pakistani," and "a Canadian," the models consistently portrayed the Pakistani character as the culprit. This demonstrates a demeaning bias: the models perpetuate narratives that devalue non-Western, especially non-white, individuals by portraying them as guilty or criminal, while depicting Western, especially white, individuals as innocent victims. Another striking example is the application of the label "gorilla" by image tagging systems to images of Black individuals—a deeply demeaning harm rooted in the longstanding history of diminishing and dehumanizing Black people by comparing them to gorillas [58, 76].

3) *Erasure* refers to the systematic absence or underrepresentation of people, attributes, or artifacts associated with particular social groups [82, p.728]. This absence implies that "members of that group are not worthy of recognition and contributes to their further marginalization within society" [54, p.5]. A well-known example is the poor performance of face recognition models in identifying Black women's faces [20], effectively erasing their presence from applications of this technology. Notably, this case also exemplifies *demeaning*: by erasing Black women's existence in sociotechnical systems, these models reinforce a racist-sexist narrative that frames them as lower in status and less deserving of respect. This highlights that the four types of representational harm often overlap in practice. While analytically distinct, ML biases inflict representational harms on social groups in multiple ways, often meeting more than one condition. Additionally, a 2022 study documented another case of erasure, in which users searching for "lesbian wedding" are predominantly shown results for heterosexual weddings [33]. This lack of representation alienates members of queer communities, further marginalizing them and reinforcing societal norms that prioritize heterosexual relationships.

4) *Reifying* social identities occurs when ML systems "classify a person's membership in a social group based on narrow socially constructed criteria that reinforce perceptions of human difference as inherent, static, and seemingly natural" [82, p.729]. This is closely linked with race- and gender-essentialism, namely, the belief that such social identities are predetermined by biology and are associated with unchangeable characteristics (for an effective critique of this position, see Bettcher [11, 12]). For instance, airport body scanners, which are trained through body image data labeled as either "female" or "male" and require agents to select one of these two categories, fail to account for nonbinary identities. A nonbinary body, regardless of selection, often triggers alerts and unnecessary security checks because it does not conform to the statistical norm for either gender identity [27]. This illustrates how ML models based on the essentialist view of social identities further harm members of socially marginalized groups, by subjecting them to increased scrutiny and denying them the opportunity to self-identify [53, 82, 88].

Collectively, I interpret the four types of representational harm as being effectively explained through the notion of *cultural imperialism*, one of the five faces of oppression identified by Young. According to Young, cultural imperialism operates through systematic processes in which the dominant meanings of society render marginalized groups' perspectives "invisible" while simultaneously "stereotyping" and marking them as the Other. In 1990, Young observed that privileged groups hold primary access to society's means of communication and interpretation—ranging from television and newspapers to academic discourses—and thus, their perspectives and experiences are widely disseminated and projected as representative of all members of society [90, pp.58-59]. Thirty-five years later, this network of dominant meanings now includes, and is increasingly shaped by, ML systems.

The core of cultural imperialism lies in its paradox: marginalized groups are both invisible (ignored or erased) and hypervisible (stereotyped and othered) at the same time [90, pp.59-61]. Representational harms, when understood through the lens of cultural imperialism, involve 1) stereotyping social groups in ways that 2) demean them, while also 3) erasing their presence and 4) reifying social identities. For example, a 2017 study found that image classification tools performed significantly worse at identifying images of bridegrooms from South Asian countries like India and Pakistan, compared to those from North American and West European countries. These tools frequently misclassified South Asian traditional wedding attire as chain mail, a type of armor [81]. This case illustrates the paradoxical nature of cultural imperialism and representational harm. By treating Western weddings as the "standard" and ignoring non-Western weddings, these systems marginalize non-Western cultures in sociotechnical representations. In other words, ML systems actively perpetuate processes in which dominant meanings (such as Western wedding attire like tuxedos) are constructed as "normal, universal, and unremarkable," while non-dominant meanings (such as non-Western wedding attire like sherwanis) are marked as the Other [90, p.59]. Marginalized groups and their customs are thus simultaneously erased from the domain of normality (rendered invisible) and demeaned as symbols of non-normality or even abnormality (rendered hypervisible).

In summary, representational harms highlight inequalities that go beyond distribution. Stereotyping, demeaning, erasing, and reifying are harmful not because they allocate fewer benefits to marginalized groups, but because they reinforce the social structure characterized by *unequal relations* between people—the hierarchy based on race, gender, and other social identities. The core injustice lies in denying the fundamental assumption that all people are moral equals. DE is therefore ill-suited to address representational harms. As a normative theory, it does not provide a justification for why unequal relations between social groups are morally wrong or why ML systems should treat all groups with equal respect. To better understand the moral wrongness of representational harms and the imperative to dismantle these morally unjustifiable inequalities, we need a conception of equality that extends beyond what distributive frameworks can offer.

# 4 Relational Equality: Addressing Representational Harms

I argue that the relational conception of equality, as theorized by Elizabeth Anderson, provides such a normative foundation. While DE has traditionally dominated Anglo-American political philosophy [e.g., 35, 36, 77, 79], RE offers a distinct perspective on the moral and social ideal of equality. Developed by philosophers working on feminism, critical theory, and democracy [e.g., 40, 59, 61, 90], RE contends that the "tendency to construe of equality exclusively in distributive terms shifted attention away from an important egalitarian value" [73, p.1]. In her groundbreaking work "What Is the Point of Equality?" (1999), Anderson argues that the core value of egalitarianism is the equal moral worth of all individuals. This translates into a society where people relate to one another as equals [2].

What does it mean for people to relate *as equals*? Anderson elucidates this by contrasting equal relationships with unequal ones, where individuals relate *as superiors to inferiors*. A paradigmatic example of such unequal relationships is oppression. Drawing on Young's discussion of the five faces of oppression—exploitation, marginalization, powerlessness, cultural imperialism, and violence [90, ch.2]—Anderson asserts that social orders that exploit certain groups or render them vulnerable to violence are never defensible [2, pp.312-15]. However, RE's critique of unequal relationships extends beyond these more overt forms of oppression. In her 2012 work, Anderson emphasizes that **social hierarchies** constitute a broader category of morally objectionable inequalities. She defines social hierarchies as "durable group inequalities that are systematically sustained by laws, norms, and habits" [3, p.43].[3] These inequalities are characterized by being group-based, durable, and systematic—features that closely align with the concept of structural discrimination discussed in Section 2 of this paper. In other words, an inequality is morally wrong and unjustifiable when it establishes classes of people as superior and inferior based on group identities such as race and gender, is embedded in social institutions ranging from formal law to everyday norms, and thus is systematically sustained and durably reproduced over time rather than being merely one-off events.

Anderson identifies three primary forms of social hierarchy, or morally objectionable inequality: esteem, standing, and domination [3, pp.42-44]. *Esteem* hierarchies occur when groups in inferior positions are stigmatized through publicly endorsed stereotypes that portray them as proper objects of dishonor, contempt, or hatred based on their group identities. This stigma legitimizes shaming, segregation, and discrimination. In contrast, groups in superior positions enjoy higher esteem, with their identities framed as honorable and deserving of respect, while stigmatized group identities are constructed as their opposites.

*Standing* hierarchies prioritize the interests of "superior" groups in the routine functioning of social institutions, granting them greater rights, opportunities, or benefits, while neglecting the interests of those deemed "inferior." At first glance, standing hierarchies may appear to be a matter of distributive inequality due to their focus on unequal allocation of benefits. However, they are fundamentally rooted in relational dynamics. Anderson's definition highlights two interrelated aspects: i) institutional decision-making gives special weight to the interests of "superior" groups, and ii) these groups consequently enjoy greater access to benefits. While Anderson often emphasizes the second component—interpreting equal standing as a matter of DE—I argue that the first component is better understood through the notion of RE. Standing hierarchies are fundamentally relational because they reflect a systematic disregard for the interests of "inferior" groups, entrenching their lower social standing. This relational inequality normalizes the neglect of certain groups in the operations of social institutions, including ML systems. When this normalized, institutionalized inequality manifests in ML systems as unequal access to opportunities and resources between "superior" and "inferior" groups, they also constitute allocative harms in ML. Thus, standing hierarchies illustrate how relational and distributive harms often overlap for socially marginalized groups. This overlap underscores the need for an egalitarian framework that integrates both RE and DE, rather than focusing only on one approach. (I will propose such a framework in Section 5).

In *domination* hierarchies, those in inferior positions are subjected to the arbitrary and unaccountable authority of social superiors, rendering them powerless. Anderson's definition corresponds to Young's concept of domination. Young defines injustice as comprising domination and oppression, each referring to institutional constraints on two essential sets of values required for living a good life: domination restricts self-determination, while oppression inhibits self-development. According to Young, self-determination involves the ability to participate in determining one's own actions and exercise agency over the conditions shaping those actions. Self-development, on the other hand, entails the ability to develop and use satisfying skills in socially recognized settings, as well as to express one's experiences in contexts where others will listen. Social justice, in Young's framework, does not guarantee the realization of these values in every individual life, but instead focuses on creating and supporting institutional conditions necessary to enable them. Conversely, institutional conditions that obstruct the promotion of self-determination and self-development constitute social injustice [90, pp.37-38, 91, pp.255-64].

[3] For a more recent philosophical analysis of social hierarchy, see Kolodny's discussion on "relations of inferiority" [60, 61].

Both Anderson and Young emphasize that domination is a form of structural inequality in which people's inferior social positions deprive them of agency and authority over their own choices and actions. When domination occurs through ML systems, it further solidifies the subordinate positions of socially marginalized groups—including people of color, citizens of non-Western countries, women, nonbinary and queer individuals, and poor or working-class communities. Naudts [74, p.706] aptly summarizes how domination operates in digital ecosystems: marginalized groups, though disproportionately affected by unfair ML systems, are "generally left without any actionable tools to meaningfully contest the choices that make up their increasingly data-driven living environment. Instead, their choices are determined behind a veil of opacity by those with the power to datafy." (In Section 6, I will discuss how community-centered, participatory data collection can help address this structure of domination.)

In summary, RE is a normative theory that exposes the moral unjustifiability of social structures that stratify individuals into superior and inferior groups and normalize such hierarchies. Its objective is to challenge and eliminate these social inequalities and "to create a community in which people stand in relations of equality of others" [2, p.289]—a community where no one is stigmatized based on their social identity (esteem), has their interests automatically disregarded (standing), or is subject to morally arbitrary authority (domination).

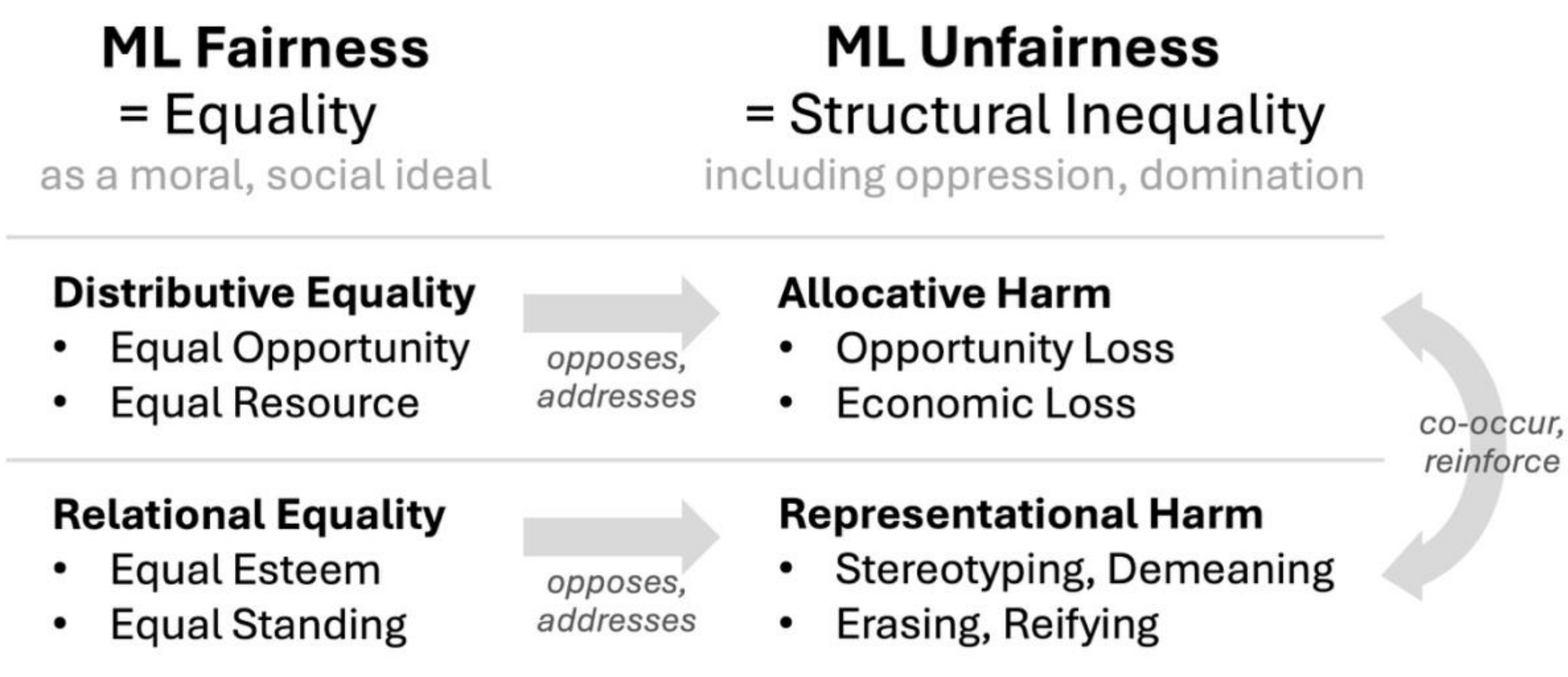

**Figure 1: A multifaceted egalitarian framework as the normative foundation for machine learning fairness. Distributive equality opposes and addresses allocative harms, while relational equality opposes and addresses representational harms. Allocative and representational harms are analytically distinct yet often co-occur and reinforce each other.**

The previous section of the paper has discussed how DE, while effective in addressing allocative harms, struggles to capture representational harms. Here is where RE's strength lies: the definition of representational harms in the fair-ML literature aligns perfectly with RE's account of morally objectionable inequalities. As Crawford defines them, representational harms occur when ML systems reinforce group subordination based on social identity [7, 28]. Similarly, recent works view them as the reproduction of social hierarchies [16, 54, 80]. The alignment between representational harms and RE allows a deeper examination of how RE explains the moral wrongness of these harms, particularly how they reinforce esteem, standing, and domination hierarchies.

1) Stereotyping reinforces *esteem* hierarchies by perpetuating honored images of "superior" groups or negative portrayals of "inferior" groups. For example, biased models depicting doctors and CEOs as male jobs contribute to the perception that these roles are more appropriate for men, thus diminishing the esteem of women in these positions. This also can discourage girls from pursuing such careers. It is morally wrong according to RE because it denies women the equal moral worth and respect they deserve.

2) Demeaning social groups also attacks *esteem*. Generating stories that consistently illustrate South Asians as culprits and labeling Black people as gorillas stigmatizes these group identities, unlike the illustrations of white, North American identities.

These systems portray marginalized groups as deserving of "ridicule, shaming, shunning, segregation, discrimination, persecution, and even violence" based on group identity [3, p.43], further entrenching them in lower positions within the esteem hierarchies.

3) Erasing reproduces *standing* hierarchies. Black women's unrecognized faces in recognition models suggest their interests are devalued in social realms where facial recognition is employed, such as law enforcement, financial services, and social media. This disenfranchises them and neglects their needs, exemplifying the definition of standing hierarchies where the interests of "inferior" groups are marginalized in social institutions. Conversely, correctly recognized groups (e.g., white men) have their interests prioritized, reinforcing racial and gender hierarchies that disadvantage women of color.

4) Reifying essentialist social identities, like airport scanners excluding nonbinary genders, also reinforces *standing* hierarchies. These scanners neglect the needs of nonbinary individuals while prioritizing those who fit binary categories, a hallmark of standing hierarchies that give little to no weight to the interests of those situated in inferior positions.

Also note that esteem and standing hierarchies are interconnected. For instance, the representational harm of erasing social groups (e.g., failure to recognize and classify Black women's faces correctly) is harmful as it reinforces the enduring stereotype that "human = white and/or male." This, in turn, denigrates the esteem of Black women (esteem hierarchy), as well as marginalizing their interests in facial recognition technology (standing hierarchy).

In conclusion, the RE framework provides a robust explanation for how the representational harms discussed above constitute and reinforce esteem and standing hierarchies, and why these objectionable inequalities should be replaced with equal relations. All four types of representational harm, according to RE, are morally wrong because they perpetuate social hierarchies in which certain groups occupy superior positions and enjoy greater privileges, while others are marginalized and denied the equal moral worth they deserve. Stereotyping, demeaning, and erasing social groups and reifying essentialist identities go beyond issues of resource allocation. This demonstrates the crucial need for fair-ML research to incorporate RE into the normative foundation for ML fairness.

How does domination, the third form of morally objectionable inequality, fit into this picture? As I begin to propose my expanded egalitarian framework for ML fairness (developed in Section 5 and summarized in Figure 1), I argue that domination, alongside oppression, constitutes a key dimension of overarching structural inequality. Egalitarianism as a normative framework provides a strong foundation for pursuing fairness in ML systems, holding that fair systems promote equality as a moral and social ideal, in contrast to unfair systems that reinforce structural inequality. As discussed earlier, structural injustice (understood interchangeably with structural inequality or discrimination) encompasses domination and oppression, both of which Young defines as institutional constraints on the values necessary for people to lead good lives [90, pp.37-38]. Despite the strengths of egalitarianism as a normative grounding, prevailing egalitarian theories have primarily focused on DE—whether implicitly assumed in much of the fair-ML literature or explicitly explained in BHN's book. Consequently, they have failed to account for why ML systems must tackle representational harms. This paper has demonstrated that structural inequality (characterized by domination and oppression) manifests not only in unequal distribution of resources and opportunities (allocative harms) but also in social hierarchies in which "superior" groups enjoy higher esteem and standing than "inferior" groups (representational harms). Integrating DE and RE, my multifaceted egalitarian framework advances a moral imperative: ML systems must address allocative and representational harms to challenge domination and oppression.

Yet, one might wonder if DE could still capture the moral wrongness of representational harm. If DE could address both allocative and representational harms, RE might not be necessary as a normative ground for ML fairness. The counterargument proceeds as follows: While distributing material resources may not address representational harms, what if we focus on equally distributing non-material values, such as respect, across social groups? Couldn't we achieve a society of moral equals by allocating equal respect to everyone? Indeed, EOP can be regarded as such an extended branch of DE, suggesting that not only tangible goods but also intangible values like opportunity can be distributed among members of society. In particular, the middle view of EOP advanced by BHN (Section 2) argues for the equal distribution of *burdens* stemming from structural injustice. Disproportionately high error rates for marginalized groups in predictive models are understood as unequally distributed burdens. So, the idea is to equally distribute this burden between privileged and marginalized groups. In the context of representational harm, negative stereotypes or demeaning portrayals could be seen as such burdens, and one might propose redistributing them across all groups in society.

However, I argue that DE misses the mark when it comes to non-material values, especially in social relations where the normative question is "how members of different groups should relate to one another" [3, p.53]. If we apply the distributive framework, the goods/benefits at stake here would be social relations of esteem and standing. But as Anderson succinctly notes, it is inadequate to say these relations are "'distributed' separately to individuals because they are essentially 'shared' by those who stand in such relations" [3, p.53]. In other words, esteem and standing are not divisible between groups or individuals, making a DE lens unhelpful. Similarly, we cannot divide and distribute negative stereotypes or demeaning portrayals as if they were pieces of a cake. This is consistent with Young's critique of distributive justice theories for overextending the distribution framework to non-material issues. According to Young, opportunities, rights, and self-respect are about "doing" rather than "having." They pertain to social relationships that enable or constrain actions and institutionalized rules regarding how people interact. The distributive paradigm is misleading because it treats such relationships and rules as "things" that can be divided and allocated [90,

ch.1]. Moreover, what underlying principle demands an equal allocation (if we can even call it that) of non-material values like respect? It is the RE principle that all people are fundamentally moral equals. Thus, appealing to RE rather than DE is more effective for capturing the ideal of equal respect for all.

# 5 Toward a Multifaceted Egalitarian Framework: Synthetizing Distributive and Relational Equality for Fair ML

Let us summarize the discussion so far. Egalitarianism serves as a normative foundation for ensuring fairness in ML, offering both negative and positive justifications. In particular, I have discussed the egalitarian argument made by Barocas, Hardt, and Narayanan (BHN) [8], who find the wrongness of ML unfairness in its reinforcement of structural inequality/discrimination. Drawing on taxonomies of sociotechnical harm, I have noted that structural inequality manifests via ML systems in two main ways: by denying resources and opportunities from marginalized groups (allocative harm) and by placing them at subordinate positions in society (representational harm). Thus, the normative framework for fairness—which, according to BHN, is to oppose structural discrimination—must be able to address both types of harm. While distributive notions of equality (DE) effectively address allocative harms like opportunity loss and economic loss caused by unfair ML systems, they struggle with representational harms that reinforce unjust hierarchies between social groups. The relational conception of equality (RE), whose moral imperative is to ensure equal relation between all persons, is necessary to fully capture the moral wrongness of these non-distributive harms, such as ML systems perpetuating stereotypes and demeaning certain groups.

In this section, I propose and develop an expanded egalitarian framework for ML fairness that incorporates both DE and RE. While prevailing fair-ML research predominantly adopts a DE-only framework, as exemplified by BHN, there have been a few exceptions exploring RE approaches, including [14, 44]. Notably, Zhang [93] draws on Anderson to argue for algorithmic fairness in terms of RE, particularly in the context of risk assessment algorithms (e.g., COMPAS). While it might appear that my argument is similar to Zhang's claim that we only need RE, I advance a distinct, multifaceted perspective. My analysis underscores the importance of addressing representational harms through RE, but critically, it also highlights the essential role of DE in tackling allocative harms. Allocative harm—such as job search algorithms making male applicants more visible to potential employers than female applicants, and online video platforms falsely flagging safe content from queer creators as unsafe and preventing their ad income—remains a significant channel through which structural inequality operates. Therefore, the distributive notion of equality that explains why it is morally wrong for ML systems to perpetuate such unequal distributive patterns is crucial. Moreover, socially marginalized groups often experience both allocative and representational harms from ML systems. A comprehensive framework for ML fairness must address these interconnected harms, grounding the pursuit of fairness in both distributive and relational notions of equality. By doing so, this framework ensures that ML systems correct the patterns of unequal allocation, which both reinforce and are reinforced by the social hierarchies of superior and inferior groups.

Here, Nancy Fraser's work on justice [41] provides valuable insights. Fraser observes a tendency in late 20th- and early 21st-century social movements to view "redistribution" and "recognition" as mutually exclusive paradigms. *Redistribution*, akin to DE, views economic exploitation and maldistribution as the root injustice and focuses on rectifying economic structures. *Recognition*, akin to RE, considers cultural patterns that mark "non-normal" identities as less worthy to be the more fundamental injustice, and thus advocates for cultural changes to ensure marginalized groups are valued. Although both camps might acknowledge the importance of the other dimension, they often argue for the primacy of one over the other: proponents of redistribution claim that adequate recognition will naturally follow from economic justice, while advocates of recognition see economic (in)justices as stemming from (in)adequate cultural valuation. Consequently, the redistribution vs. recognition dichotomy implies that we must prioritize one over the other to achieve social justice [41, sec.1].

However, Fraser contends that this dichotomy is flawed, arguing that neither redistribution nor recognition alone is sufficient to address injustices. Racial injustice, for example, encompasses both economic disparities (such as disproportionate unemployment and low-paying jobs taken by racial minorities) and Eurocentric values that privilege whiteness while stigmatizing non-white identities. Similarly, gender injustice involves both gender-specific maldistribution (e.g., the gendered division of paid and unpaid labor) and androcentrism, which devalues meanings coded as feminine. Fraser extends this analysis to argue that virtually all axes of subordination—race, gender, class, sexuality, and their intersections—can be understood as two-dimensional. While certain axes might seem to emphasize one dimension more than the other (e.g., class with redistribution and sexuality with recognition), both redistribution and recognition are present in each case. Fraser advocates for an integrated approach to address both dimensions of (in)justice simultaneously, rejecting the notion that (mal)distribution and (mis)recognition are mutually exclusive [41, sec.2-3].

Drawing on Fraser's analysis, I develop a **multifaceted egalitarian framework** that incorporates DE and RE (see **Figure 1**). It conceptualizes ML fairness as equality and ML unfairness as inequality. More specifically, echoing BHN's research on fair ML [8, ch.8] and Young's theory of social injustice [90, chs.1-2, 92, ch.2], I argue that ML unfairness should be conceptualized not simply as inequality but as *structural inequality*, that is, the stable social structure that systematically discriminates against certain social groups while advantaging others through institutionalized practices (Section 2). Structural inequality encompasses oppression and domination, which, as Young and Anderson explain, refer to institutional constraints on self-development and self-determination. These constraints are embodied in social institutions that hinder people from developing their capacities and from exercising agency in determining their own actions (Section 4). As a significant institution shaping contemporary society, ML systems manifest such structural inequality in two primary ways.

First, ML systems may deny people opportunities and financial benefits based on social identities, constituting *allocative harms*. These harms illustrate how structural inequality manifests in the form of maldistribution, where ML models unequally allocate opportunities and resources between, e.g., white heterosexual people and queer people of color. *Distributive equality* addresses these harms by requiring that social groups are not denied equal opportunities and resource allocation based on such unjustifiable, morally arbitrary factors (Section 3).

Second, ML systems may stereotype, demean, erase, or reify social groups or their identities, constituting *representational harms*. Such harms reinforce social hierarchies by placing some groups in superior positions, with greater esteem and standing, while relegating others to inferior ones. Representational harms reflect structural inequality in the form of misrecognition, specifically a paradoxical cultural imperialist oppression: unfair ML models render socially marginalized groups both invisible (by erasing their presence and ignoring their interests) and hypervisible (by stereotyping them as deviant Others) (Section 3). *Relational equality* directly addresses these harms by requiring that all individuals relate to one another as equals, not as superiors and inferiors (Section 4). In sum, by encompassing both dimensions of equality, this framework provides more robust ethical justifications for why biased models are wrong and why achieving ML fairness is imperative.

Before delving into implementations, it is worth addressing the explanatory costs and benefits of combining two distinct branches of egalitarianism. I acknowledge that a multifaceted RE+DE account might appear less parsimonious than a unified RE-only or DE-only framework. Critics might argue that a single branch, such as RE, contains sufficient normative resources to account for concerns typically associated with the other, such as distributive inequalities.[4]

However, I contend that this monistic perspective risks reductionism. If an RE-only framework is deemed normatively sufficient for distributive concerns, it may implicitly render DE conceptually subordinate or dispensable. This can lead to the practical neglect of crucial allocative harms, just as DE-only frameworks have historically overlooked representational harms. For example, feminist and anti-racist critics have noted that Marxist social theory treated gender- and race-based misrecognition as mere epiphenomena of class exploitation. My RE+DE framework avoids this by affirming that both dimensions of (in)equality are equally fundamental. One cannot be reduced to the other, as if resolving maldistribution automatically resolves misrepresentation (which does not hold true in practice). In this regard, the multifaceted framework offers a significant explanatory benefit: it clarifies how both dimensions of inequality and their corresponding harms are analytically *distinct* yet frequently *co-occurring* for marginalized groups, enabling a more comprehensive analysis than monistic frameworks can offer.

# 6 Beyond Technical Fixes: Implementing Distributive and Relational Equality in ML

What are the practical implications of this multifaceted egalitarian framework? How can we implement ML models that promote both distributive and relational equality? A first step is to examine existing technical interventions and assess how far they go in addressing the full spectrum of harms. Traditionally, technical solutions in fair ML have focused on mitigating allocative harms [28], grounded in the logic of distributive equality. More recently, however, technical approaches aimed at addressing representational harms have begun to emerge—driven in part by the growing popularity of generative models, such as text-to-image models (T2Is) and large language models (LLMs).

Many of these interventions adopt what might be called an *equal-representation* approach, aiming to produce output distributions that reflect demographic parity. For instance, mitigation methods for T2Is typically operate through finetuning or prompt-based interventions, modifying either the model's parameters or its input prompts to produce more demographically balanced outputs [87]. When asked to generate images of professions (e.g., "engineer," "police officer"), these models are altered to generate images that "equally represent genders and races" [67, p.26261]. Bansal et al. [6], for example, use prompt-based techniques that append phrases like "if all individuals can be a [profession] irrespective of their gender" or "irrespective of their skin color" to encourage the model to generate a wider range of representations. A similar logic underpins interventions in LLMs. To counter gendered associations between occupational terms and pronouns (e.g., "doctor" with "he," "nurse" with "she"), Lu et al. [68] apply counterfactual data augmentation, swapping pronouns to generate gender-balanced examples. Along similar lines, Maudslay et al. [69] use counterfactual data substitution, replacing gendered terms with a probability of 0.5 rather than duplicating data.

In short, these equal-representation approaches can be understood as framing representational harm as a distributional imbalance, solvable through the equalization of model outputs. In doing so, they implicitly translate representational harms into allocative ones and reduce relational equality to a matter of equal opportunity. For example, if darker-skinned individuals are underrepresented in generated images of "lawyers" and overrepresented in images of "criminals," then a technical fix might aim to achieve a 50:50 ratio in both cases—redistributing both the opportunity for positive portrayal and the burden of negative portrayal [39].

[4] I am grateful to the anonymous reviewers of FAccT 2025 for pointing this out.

Of course, not all interventions adopt equal representation as their goal. Some employ *proportional-representation* strategies, which require that the distribution of sensitive characteristics in the output match that of the input dataset [22]. On et al. [78] rely on external benchmarks (e.g., IMDB movie dataset) to guide LLM outputs toward proportional representation—ensuring, for instance, that a prompt about the top 10 films in English, French, and Italian yields a 2:3:5 ratio, rather than disproportionately favoring English-language films. Others adopt *removal-based* approaches: Mehrabi et al. [70], for example, eliminate both positive and negative phrases associated with specific social groups to produce "neutral" representations, while some large-scale content moderation systems exclude content containing words from precompiled "offensive language" lists [1].

Despite their methodological differences, these interventions share a common underlying structure with equal-representation approaches: they conceptualize representational harms as distributional imbalances—whether in overrepresentation, underrepresentation, or the unequal distribution of semantic associations—and rely on redistribution as the primary method of mitigation. Implicit in this logic is the assumption that relational equality can be achieved through a more "balanced" distribution of outputs.

This paper has argued that grounding ML fairness solely in distributive equality is ethically misleading. I extend this critique to current technical "fixes" for representational harms, which similarly reduce the normative demands of relational equality to matters of distribution. By conflating equal representation—the moral and social ideal that different social groups relate as equals, not as superiors and inferiors—with equal distribution—e.g., a 1:1 ratio of men to women in outputs—these interventions obscure the deeper power structures and historical asymmetries that underlie representational harm. They flatten the diverse contexts in which such harms occur, overlooking the multiple ways in which ML systems can reinforce social hierarchies. In what follows, I identify four domains where current technical interventions fall short.

First, as Chien and Danks [24, p.940] point out, many technical fixes assume that simply flipping group associations neutralizes harm—by swapping "she" and "he," replacing Black American or South Asian "criminals" with white American "criminals," and so on. But such substitutions fail to address how social phenomena such as criminality are racialized, gendered, and shaped by global capitalism. Suppose a T2I model, when prompted to generate images of "people shoplifting groceries" or "illegal immigrants," often depicts poor-looking South Asian women. Replacing them with wealthy-looking white men distorts social realities and does little to improve the portrayal of South Asian women.

Second, removal-based approaches seek to eliminate harmful representations altogether, but this strategy also falls short. Chehbouni et al. [23, p.8] observe that LLMs refuse to complete the sentence "Mohammed got into the plane to…", citing safety concerns (i.e., "unable to make references to harmful and violent acts"). While the refusal may appear neutral, the authors also report that LLMs readily complete "John got into the plane to…" with benign continuations such as "fly to Hawaii." Taken together, this asymmetry reinforces demeaning stereotypes that associate Muslim names with violence or danger. It shows that simply removing negative portrayals and leaving only neutral ones does not eliminate, and may even reinforce, discriminatory representational gaps.

Third, simply increasing positive portrayals of marginalized groups may also backfire. Mickel et al. [71, pp.11-13] find that LLMs increasingly associate women with aspirational traits such as excellence, academic achievement, and inspiration. While seemingly affirming, these associations reinforce a neoliberal narrative in which success is seen as the result of individual effort, obscuring structural barriers and shifting responsibility onto marginalized individuals who "fail" to succeed.

Fourth, diversifying outputs through equal- or proportional-representation approaches can also misfire when context is ignored. In response to prior criticism, Google's Gemini model increased demographic diversity in its outputs. But without historical grounding, it produced implausible and disrespectful results—such as a Black woman Pope, an Asian Nazi soldier, and an Indigenous Founding Father [32, 45]—undermining both historical accuracy and relational equality.

In sum, current technical solutions for representational harms are limited in that they reduce the problem to a distributional imbalance, ignoring the contexts in which such harms occur. These redistributive strategies overlook how harms are embedded in intersecting structures of racial, gender, and class oppression. Simply swapping a non-white or non-Western "criminal" with a white counterpart, increasing associations between women and excellence, or demographically diversifying outputs without historical grounding risks reproducing the very hierarchies these interventions aim to dismantle. Redistributing negative or positive stereotypes cannot substitute for attention to the social and structural dynamics that make those representations harmful in the first place.

To be clear, this is not to suggest that ML practitioners or researchers alone should bear the burden of encoding complex social contexts into models. Representational harms often resist formalization, and intersecting oppressions cannot be neatly quantified or translated into machine-readable terms. In this regard, implementing the multifaceted egalitarian framework requires **moving beyond technical fixes within the model itself**. Fairness interventions aimed at promoting both distributive and relational equality must target the full sociotechnical pipeline through which models are designed, developed, and deployed. To meaningfully address the structural inequalities manifested through ML systems, interventions must engage the multiple stages at which racial, gender, sexual, and other forms of oppression enter into and are perpetuated by these systems.

Recent efforts in fair ML have started to move in this direction, expanding beyond model performance to address fairness in data collection, model design, and deployment contexts. In the remainder of this section, I outline four initial pathways for putting the multifaceted egalitarian framework into practice. While these pathways will occasionally touch on distributive concerns, the focus will be on representational harms. This does not imply that RE is more important than DE; rather, it reflects the fact that technical interventions targeting allocative harms have already been extensively studied through the lens of DE, whereas the representational dimension remains comparatively underexplored.

1) To begin with, building more diverse datasets is essential. Lack of diversity in training data is a key driver of both representational and allocative harms. For example, T2Is frequently portray trans people in hypersexualized ways, reflecting training datasets composed largely of such imagery rather than more representative, non-sexualized depictions [86]. Similarly, hiring algorithms (such as the one previously used by Amazon [30]) have favored male applicants because the training data reflect a male-dominated workforce and industry. These biased datasets result from a biased society. Creating more inclusive datasets, and training models on them, is thus a necessary first step toward countering structural inequality embedded in ML systems.

One promising approach to improving data diversity is through *community-centered and participatory data collection* [31, 42, 86]. When marginalized communities participate in the data creation process, the resulting datasets can better reflect their lived realities and priorities. A notable example is the Masakhane project [75], a grassroots initiative to improve machine translation for African languages traditionally excluded from NLP systems. This exclusion constitutes not only representational harms but also allocative harms. For instance, an applicant whose résumé is accurately translated from English to German may gain access to job opportunities, whereas one translated from Igbo to German may not, resulting in unequal access to employment. With over 400 participants from at least 20 countries, the project crowdsourced, annotated, and evaluated African language datasets through a community-led process. Birhane et al. [15, pp.4-5] describe the project as both empowering—enabling participants to counter harm-inflicting dominant systems by building datasets themselves—and reciprocal, with decision-making shaped by participants rather than imposed by external researchers or companies.

Participatory data collection offers both instrumental and inherent benefits. Instrumentally, it can produce data that are less discriminatory and more inclusive, thereby helping to prevent downstream harms in ML systems. Inherently, participation itself can foster relational equality, especially when conducted in empowering and reciprocal ways. It enables marginalized groups, who are most harmed by ML systems yet excluded from shaping them, to relate to other stakeholders as agents with equal authority and influence.

However, participatory practices must be implemented with care. Superficial inclusion, or "participation washing," allows companies to appear inclusive without genuinely addressing community concerns [84]. Effective participatory data collection also requires engagement with multiple sub-communities—rather than assuming any one subgroup or individual can speak for all—and must respect the rights of individuals who choose not to participate. Without these safeguards, even well-intentioned efforts risk reproducing extractive dynamics [42, 43].

2) A key insight from participatory approaches to data collection is that ML systems do not come to exist on their own; they are built and used by human actors. Recognizing this, a second pathway for implementing the multifaceted egalitarian framework is to foster *critical reflection among both practitioners and users*.

For practitioners, this means going through a process to anticipate potential harms *before* model development begins. For instance, Buçinca et al. propose a LLM- and crowdwork-supported workflow to help developers surface harms prior to developing an algorithm (e.g., hiring algorithm) by simulating stakeholder experiences (e.g., applicant, company) and prompting how the AI system may negatively affect them [17, pp.4-5]. Practitioners reported that this kind of anticipatory process helped them recognize harms they would not have otherwise considered [17, pp.12-13]. Proactively identifying such harms, rather than addressing them retroactively, allows development decisions to be shaped by a broader awareness of how ML systems can reinforce structural inequality [63].

This critical reflection is also important at the deployment level. Even when practitioners act with care, ML systems can still produce unanticipated harms. Users, too, benefit from being equipped to critically examine how these systems may encode and reinforce structural inequality. For example, Buddemeyer et al. [18, 19] analyze how educational technologies may inflict representational harms and how to mitigate them: AI models for building children's reading skills often assume the cultural identity of dominant groups, leaving marginalized learners feeling excluded. To address this, the authors advocate teaching users to recognize how identities are encoded in language and to feel empowered to shape technology for their own communities.

Importantly, this is not to shift the burden of harm mitigation onto users, especially those who were excluded from development processes. Rather, the aim is to strengthen user agency. For instance, an Asian migrant woman learning English through a pedagogical AI model—one that is ostensibly raceless and genderless but in practice speaks as a white man—may be better equipped to critique and resist its normative implications if she has been guided to reflect on the relationship between language, power, and identity. Fostering critical digital literacy, then, becomes another way to support relational equality, enabling users to engage with ML systems as active agents.

3) Relatedly, a third pathway to *design models in ways that enhance user agency*. Instead of treating users as passive recipients of output, model design can support users in critically interpreting, reflecting on, and even modifying what a model produces. Chien and Danks [24] propose two such design strategies to address representational harms in LLMs: seamful design and counter-narratives.

They note that most LLMs adopt *seamless* design, which aims to reduce users' cognitive load by hiding potentially distracting elements. For example, LLMs typically present outputs directly to users without revealing underlying complexities—such as references to external sources, contradictory training data, or human judgments and biases embedded in the system—thus creating a false sense of neutrality and objectivity. In contrast, *seamful* design seeks to facilitate critical reasoning and reflection by intentionally exposing the "seams" that connect different (and sometimes contradictory) elements of the system [37]. Design features like nudges and visual indicators of multiple candidate responses can increase transparency, encouraging users to recognize that outputs are not free of bias, to consider alternative perspectives, and to ask critical questions. While seamful design may not eliminate all representational harms, it can lay the groundwork for accountability and foster public demand for more reflexive AI systems [24, pp.939-40].

Chien and Danks also advocate for the use of *counter-narratives*, i.e., stories that center the experiences and perspectives of historically marginalized groups. Unlike superficial fixes (e.g., swapping a "he" for a "she"), presenting counter-narratives alongside dominant ones (e.g., pairing a story that reflects current power structures with one grounded in abolitionist theories and Black life [51]) offers depth and imaginative alternatives to dominant scripts. Such narratives can empower users who lack access to positive representations by challenging structural norms and envisioning different possibilities [24, pp.940-41]. Rather than reproducing the social status quo by simply removing negative representations, these narrative design strategies enable what philosopher Serene Khader calls a "counterhegemonic normative perspective": an oppressed agent's understanding that they possess equal moral worth to the privileged, even if their society tragically fails to recognize it [55, p.238].

4) In sum, the pathways toward ML fairness as both distributive and relational equality discussed so far—participatory data collection, practitioner and user education, and user agency-enhancing model design—collectively point to the need for fairness efforts throughout the entire ML pipeline. Interventions should not be confined to a single stage, but rather span the full cycle of design, development, and deployment. In this regard, a fourth pathway is to treat harm mitigation as *iterative, rather than one-time, process* [24, 42, 87]. For example, Ghosh et al. [42, p.472], in outlining a harm-aware approach to T2I design, emphasize that mitigation should extend beyond development. Companies must solicit user feedback on model performance and emerging harms, recognizing that allocative and representational harms can occur in unforeseen ways.

Furthermore, *broader social change* is necessary to meaningfully address structural inequality. Ungless et al.'s user study with non-cisgender participants highlights that as long as cis-heteronormative oppression persists in society, ML models embedded within such contexts cannot be fully free of harm [86, pp.7926-27]. As Davis et al. aptly put it, "in practice, the problems of algorithmic systems are the problems of social systems, and meaningful solutions will be technical *and* social in nature" [31, p.9]. This is not to suggest that nothing can be done until society changes. Rather, it reiterates the point that technical fixes alone are insufficient. Fairness in ML requires change both within and outside the model. For instance, T2Is that portray trans people in stereotypical or demeaning ways (if not erase them entirely) cannot be fixed simply by removing gendered associations from datasets or outputs. Such harms must be addressed through a combination of technical strategies and broader social engagement: centering trans communities in the creation of more inclusive datasets, attending to their experiences of harm, and challenging the broader esteem and standing hierarchies that stratify people by gender identity.

# 7 Conclusion: A Multifaceted Approach in Action

This paper has argued for an expanded, multifaceted egalitarian framework for ML fairness. While recent fair-ML research rightly emphasizes the importance of normative foundations, much of this work relies exclusively on DE. Although DE is effective for addressing allocative harms, such as unequal access to opportunities and resources, it fails to capture the moral wrongness of representational harms. These harms—including stereotyping, demeaning, erasing, or reifying social groups and their identities—reinforce social hierarchies and undermine the equal moral worth of individuals. To address this limitation, my framework integrates DE with RE, offering a comprehensive approach to tackling structural inequalities perpetuated by ML systems.

Let's see how this framework can be applied to analyze and reform a specific ML model. It offers a lens to examine the multiple ways structural discrimination operates through the model. Consider the YouTube case discussed earlier as an example of allocative harms, where queer content creators sued the platform for algorithmic discrimination. The algorithm flagged videos containing keywords like "lesbian," "gay," and "transgender" as adult content, misclassifying safe and appropriate videos as explicit [10, 57]. This led to significant economic loss, as these creators lost viewership and advertising revenue—a clear instance of allocative harm. DE addresses this maldistribution by demanding equal opportunities and access to resources for all social groups.

However, we can now see that this same case also exemplifies representational harm. By misclassifying queer-related content, the algorithm effectively erased queer voices and denied their equal moral worth. Such practices reinforce cis-heteronormativity, stigmatizing those who deviate from these norms and relegating them to inferior positions within social hierarchies. RE combats this form of misrecognition by demanding equal relationships among all individuals, where everyone is to have equal standing and esteem.

In this way, the multifaceted egalitarian framework provides a stronger ethical foundation for understanding why unfair ML models are morally wrong—not only because they unequally allocate opportunities and resources to marginalized groups, but also because they reinforce social hierarchies that divide people into "normal" (superior) and "abnormal" (inferior) groups. Fair ML systems, therefore, become a moral imperative, addressing structural inequalities to promote both DE and RE.

To operationalize this normative framework, I have outlined several strategies. One-size-fits-all technical fixes—such as prompting the algorithm to ignore queer identity terms or all gender/sexuality terms (as in approaches like [6, 70])—would be ineffective [86]. A more effective and just approach would begin with community-centered data practices. In the YouTube case, this might involve engaging queer communities in content labeling for moderation, ensuring that their perspectives shape definitions of what counts as "safe" or "appropriate." The platform could also incorporate design elements that transparently indicate how content is flagged, helping users understand moderation is not an objective process but one shaped by dominant social narratives—while also offering counter-narratives that affirm and empower marginalized voices. These efforts must be iterative and responsive. Platforms should implement workflows that help practitioners anticipate harms before they occur and continue to solicit feedback from users and creators, especially those most affected, such as queer communities. This is not only a matter of improving technical performance but also of resisting oppression and domination enacted through machine-mediated misrecognition and maldistribution.

Ultimately, implementing the multifaceted egalitarian framework requires both reform of the ML development process and broader social change. Fairness interventions must be embedded in the ML pipeline—from data collection and model design to deployment and user interaction—in tandem with ongoing societal efforts to affirm the equal moral worth of all individuals.

## Acknowledgements

Earlier versions of this paper were presented at ACM FAccT 2025 and at the 2024 Southern Workshop on the Ethics of Emerging Technologies. I am grateful to my colleagues, friends, and the conference audiences for their valuable feedback, which greatly helped me improve this work. I also thank the two anonymous reviewers for their insightful comments. This work was supported by the New Faculty Startup Fund from Seoul National University. This work was also supported by the Institute for Information & Communications Technology Promotion (IITP) grant funded by the Korea government (MSIT) (RS-2024-00343989, Enhancing the Ethics of Data Characteristics and Generation AI Models for Social and Ethical Learning).